\documentclass[conference]{IEEEtran}
\IEEEoverridecommandlockouts

\usepackage{cite}
\usepackage{amsmath,amssymb,amsfonts}
\usepackage{hyperref}
\usepackage{algorithmic}
\usepackage{graphicx}
\usepackage{textcomp}
\usepackage{xcolor}
\usepackage{times} 
\usepackage{amsmath} 
\usepackage{amssymb}  
\usepackage{multirow}
\usepackage{multicol}
\usepackage{url}
\usepackage{afterpage}
\usepackage{stfloats}
\usepackage{cite}
\usepackage{arydshln}
\usepackage{epsfig} 
\usepackage{mathptmx}
\usepackage{subcaption}
\def\BibTeX{{\rm B\kern-.05em{\sc i\kern-.025em b}\kern-.08em
    T\kern-.1667em\lower.7ex\hbox{E}\kern-.125emX}}
\begin{document}

\title{LiRCDepth: Lightweight Radar-Camera Depth Estimation via Knowledge Distillation and Uncertainty Guidance
}

\author{
    Huawei Sun$^{1,2}$\qquad Nastassia Vysotskaya$^{2}$\qquad Tobias Sukianto$^{2}$\qquad Hao Feng$^{1}$ \\ Julius Ott$^{1,2}$ \qquad  Xiangyuan Peng$^{1,2}$\qquad   Lorenzo Servadei$^{1}$\qquad Robert Wille$^{1}$\\
    \\
    $^{1}$ Technical University of Munich, Munich, Germany \\
    $^{2}$ Infineon Technologies AG, Neubiberg, Germany\\
    E-mail: \{huawei.sun, nastassia.vysotskaya, tobias.sukianto, julius.ott\}@infineon.com\\
\{hao.feng, xiangyuan.peng, lorenzo.servadei, robert.wille\}@tum.de
}

\maketitle

\begin{abstract}
Recently, radar-camera fusion algorithms have gained significant attention as radar sensors provide geometric information that complements the limitations of cameras. However, most existing radar-camera depth estimation algorithms focus solely on improving performance, often neglecting computational efficiency. To address this gap, we propose LiRCDepth, a lightweight radar-camera depth estimation model. We incorporate knowledge distillation to enhance the training process, transferring critical information from a complex teacher model to our lightweight student model in three key domains. Firstly, low-level and high-level features are transferred by incorporating pixel-wise and pair-wise distillation. Additionally, we introduce an uncertainty-aware inter-depth distillation loss to refine intermediate depth maps during decoding. Leveraging our proposed knowledge distillation scheme, the lightweight model achieves a 6.6\% improvement in MAE on the nuScenes dataset compared to the model trained 
without distillation. Code: \url{https://github.com/harborsarah/LiRCDepth}

\end{abstract}


\section{Introduction}
Depth estimation has gained significant attention in autonomous driving as it is crucial for understanding the surroundings of the ego-vehicle. With the rapid advancements in deep learning, learning-based monocular depth estimation algorithms have been improved substantially over the years~\cite{bts,binsformer,p3depth,vadepth,dorn,lapdepth,cspn}. However, their performance is still constrained by the lack of geometric information when relying solely on camera sensors. In contrast, radar sensors provide positional data, velocities, and additional target properties. Moreover, radar offers the advantages of being low-cost and greater robustness under adverse weather conditions, making the radar-camera depth estimation task particularly appealing~\cite{lin2020depth,rc-pda,dorn_radar,radarnet,cafnet,li2023sparse}.

Despite these advantages, radar point clouds are sparser and noisier than those from LiDAR~\cite{nuscenes}. Specifically, due to the absence of height information, the 3D position of radar points is less accurate. 
Thus, recent studies~\cite{dorn_radar,mcafnet,enhance,radarnet,rc-pda,cafnet,getup} focus on processing the radar data to address the sparsity and noise, increasing the model complexity. 
Although model performance has improved by those methods, they overlook considerations of model size and evaluation efficiency. For instance, most methods utilize ResNet~\cite{resnet} to extract radar and image features, which are computationally heavy compared to lightweight alternatives like MobileNet~\cite{mobilenetv2}. However, using a lightweight backbone has drawbacks - fewer parameters limit the network's ability to fully comprehend the input, leading to reduced performance on the final task. Consequently, reducing model complexity while maintaining performance has become a crucial research area.

Knowledge distillation is a renowned technique that tackles this issue, transferring knowledge from a complex teacher network to an efficient student model. This concept was first introduced for image classification~\cite{kd_ckass1,kd_class2,kd_class3,kd_class4}. Since then, the idea of distillation has been extended to various tasks, including object detection~\cite{kd_od1,kd_od2,kd_od3}, semantic segmentation~\cite{kd_seg2,liu2019structured,kd_seg3}, and depth estimation~\cite{kd_depth1,kd_depth2,kd_depth3}. 

In multi-modal applications, it is crucial to determine which aspects of the teacher’s knowledge are essential to transfer to the student. In this work, we propose a lightweight model, LiRCDepth, with 80\% fewer parameters compared to the teacher model~\cite{cafnet}. To maintain performance, we distill knowledge in three key areas: low-level single-modal features, high-level decoding features, and inter-depth maps. The single-modal features are distilled using a conventional pixel-wise distance loss, while the decoding features are guided by pairwise structure-based similarity. For depth information, we introduce an uncertainty-aware inter-depth distillation loss, where uncertainty acts as a weighting factor during loss calculation. With these distillation methods, our lightweight student model achieves a significant performance improvement compared to training without distillation.
Additionally, to enhance final depth generation, we propose an Uncertainty-Rectified Depth Loss (URDL), which increases depth estimation accuracy. 
To the best of our knowledge, this is the first work to apply knowledge distillation to the radar-camera depth estimation task. 
Our main contributions are as follows:
\begin{itemize}
    \item We introduce an uncertainty-rectified depth loss function for network training.
    \item We design a lightweight radar-camera depth estimation framework. To ensure satisfactory performance, we distill three key components from the teacher to the student model, resulting in a 6.6\% performance improvement compared to direct training.
    \item Our model is evaluated on nuScenes~\cite{nuscenes}, achieving results comparable to other heavy-weight algorithms.
\end{itemize}

\label{sec:intro}


\vspace{-1mm}
\section{Methodology}
\begin{figure*}[ht]
\vspace{0.18cm}
\centering
\includegraphics[width=0.94\textwidth]{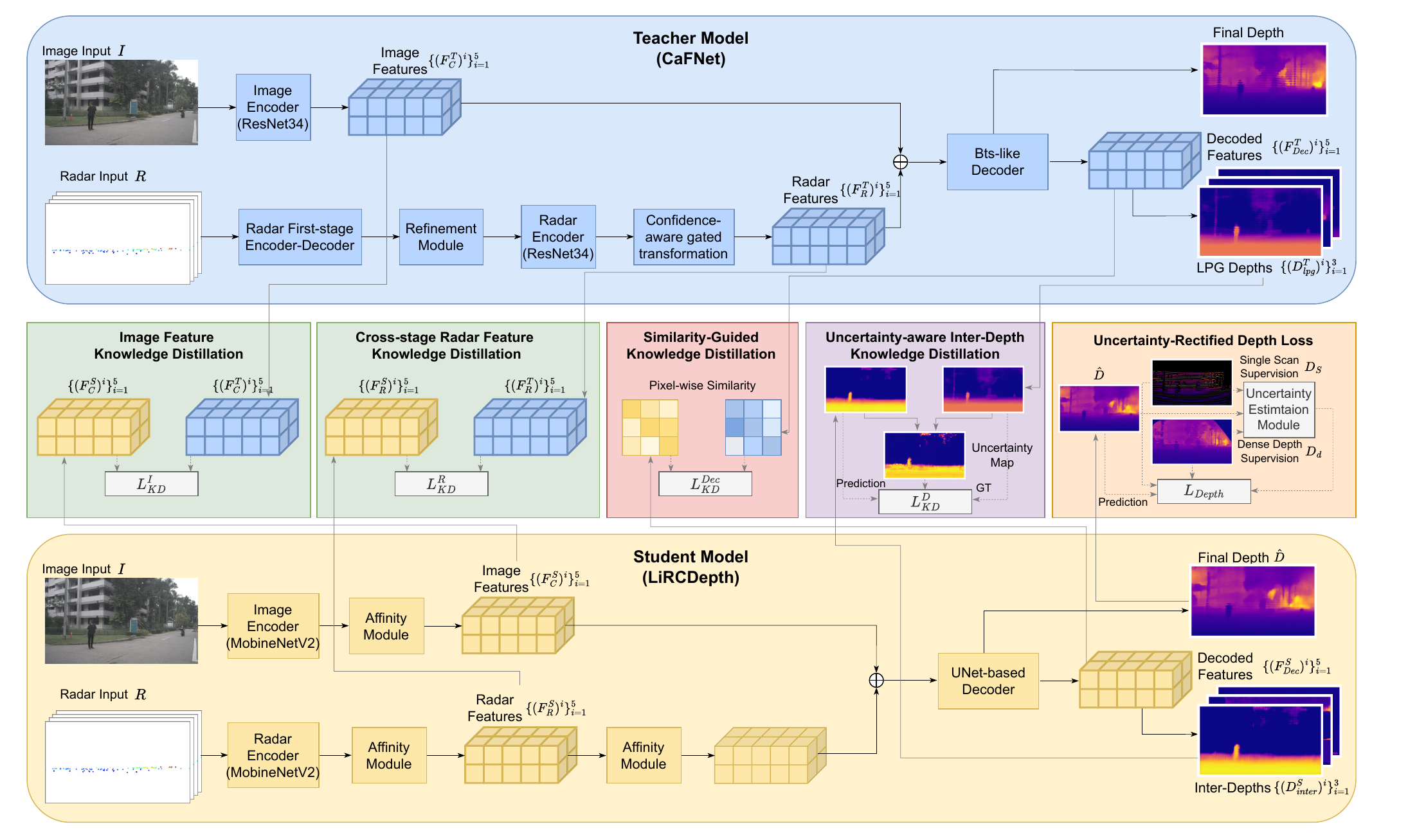}

\vspace{-2.5mm}
   \caption{Model Architecture.}
\label{fig:model}
\vspace{-6mm}
\end{figure*}

This section first introduces the model architecture. Then, we describe the knowledge distillation strategies for guiding the student network. Finally, we present the proposed URDL utilized for the depth estimation task. 
Here, the subscripts \textit{S/T/C/R} denote \textit{Student, Teacher, Camera, Radar}, respectively.
\vspace{-1.5mm}
\subsection{Model Architecture}
As illustrated in Fig. \ref{fig:model}, our approach employs the powerful CaFNet~\cite{cafnet} as the teacher model to guide the student LiRCDepth.
In the teacher model, RGB images $\mathbf{I}$ are encoded using a ResNet34~\cite{resnet} backbone, generating five feature sets $\{(\mathbf{F}_{\text{C}}^{\text{T}})^{i}\}_{i=1}^{5}$ at different scales $\frac{1}{2^{i}}$ with respect to the original image size.
Simultaneously, the radar input $\mathbf{R}$ undergoes a two-stage processing pipeline. In the first stage, $\mathbf{R}$ is refined through a predicted confidence map and a coarse depth map, followed by processing through a second radar encoder. During the decoding phase, radar features are rectified through a Confidence-aware Gated transformation Block (CaGB), and then added to the image features. The BTS-like decoder~\cite{bts} subsequently outputs the final estimated depth map along with intermediate LPG depth maps $\{(\mathbf{D}_{\text{lpg}}^{\text{T}})^{i}\}_{i=1}^{3}$.

Conversely, in the LiRCDepth, radar maps and images are encoded using MobileNetV2~\cite{mobilenetv2}. Due to the inconsistent channel length of the student and teacher features, we propose an affinity module consisting of a convolutional layer and a ReLU, enabling the channel size of the student's features to be the same as the teacher's. During decoding, the image and radar features are combined and processed by a UNet-based~\cite{unet} decoder, employing inverted residual blocks~\cite{mobilenetv2} to minimize parameter size. The decoder then outputs the predicted depth map $\hat{\mathbf{D}}$ and the inter-depth maps $\{(\mathbf{D}_{\text{inter}}^{\text{S}})^{i}\}_{i=1}^{3}$. These inter-depth maps are designed to distill key knowledge from the BTS decoder by learning the LPG depths.

To maximize the effectiveness of knowledge distillation, we distilled four specific components from the teacher model to the student model. The rationale behind selecting these components is discussed in detail in the following sections.
\vspace{-1.5mm}
\subsection{Single-Modal Feature Distillation}
This work goes beyond distilling knowledge from already fused features by also distilling single-modal features, each for specific reasons. First, the MobileNetV2 encoder is less powerful than the ResNet-34 used in the teacher model. However, the radar branch of CaFNet is more complex, making it crucial to select the most informative radar features for the student to learn from. Here, we focus on learning from the radar features $\{(\mathbf{F}_{\text{R}}^{\text{T}})^{i}\}_{i=1}^{5}$ that have been refined through the CaGB in the teacher model, as indicated in Fig. \ref{fig:model}. To achieve this, our proposed single-modal distillation loss minimizes the discrepancy between the teacher and student features.
\vspace{-1.5mm}
\begin{equation}
\vspace{-1.5mm}
\begin{split}
    L_{\text{KD}}^{\text{S-M}} &= L_{\text{KD}}^{\text{I}}+ L_{\text{KD}}^{\text{R}} \\
    &= \sum_{i=1}^{5} \frac{1}{2^{i}}||(\mathbf{F}_{\text{C}}^{\text{S}})^{i} - (\mathbf{F}_{\text{C}}^{\text{T}})^{i}||_{1} + \sum_{i=1}^{5} \frac{1}{2^{i}}||(\mathbf{F}_{\text{R}}^{\text{S}})^{i} - (\mathbf{F}_{\text{R}}^{\text{T}})^{i}||_{1},
    \end{split}
\end{equation}

\subsection{Structure-Guided Feature Distillation}
In the teacher model, a BTS-like decoder~\cite{bts} is employed during decoding, utilizing local planar guidance layers to enhance performance, while LiRCDepth adopts a relatively simple UNet-based decoder. Therefore, it is crucial that the student model's decoding features, $\{(\mathbf{F}_{\text{Dec}}^{\text{S}})^{i}\}_{i=1}^{5}$, learn from the teacher model.

Rather than using straightforward pixel-wise feature distillation, we implement an enhanced structure-guided distillation to learn high-level information from the teacher. Inspired by~\cite{liu2019structured}, we first compute pairwise similarity for each feature from both CaFNet's and LiRCDepth's decoders across five scales. 
For instance, let $(\mathbf{F}_{\text{Dec}}^{\text{S}})^{i}$ and $(\mathbf{F}_{\text{Dec}}^{\text{T}})^{i}$ represent the flattened decoding features from the student and teacher models, respectively. The pairwise similarity scores 
\vspace{-1.5mm}
\begin{equation}
\vspace{-1.5mm}
    \alpha_{p,q} = \frac{\mathbf{f}_{p}^{\top}\mathbf{f}_{q}}{||\mathbf{f}_{p}||_{2}||\mathbf{f}_{q}||_{2}},
\end{equation}
between the $p^{\text{th}}$ and $q^{\text{th}}$ pixels are calculated for both the student $(\alpha_{p,q}^{\text{S}})^{i}$ and the teacher 
 $(\alpha_{p,q}^{\text{T}})^{i}$ model.
Here, $\mathbf{f}_{p}$ and $\mathbf{f}_{q}$ represent the feature vectors of the $p^{\text{th}}$ and $q^{\text{th}}$ pixels.
To minimize the difference between the similarity maps from the student and teacher models, our loss 
\vspace{-1.5mm}
\begin{equation}
\vspace{-1.5mm}
    L_{\text{KD}}^{\text{Dec}} =  \sum_{i=1}^{5} \frac{1}{2^{i}}\frac{1}{(W^{i}H^{i})^{2}}\sum_{p=1}^{W^{i} H^{i}}\sum_{q=1}^{W^{i}H^{i}}((\alpha_{p,q}^{\text{S}})^{i}-(\alpha_{p,q}^{\text{T}})^{i})^{2},
\end{equation}  
adopts the L2 distance. Here, $H^{i}$ and $W^{i}$ represent the height and width of the feature map at the $i^{\text{th}}$ scale.

\vspace{-0.5mm}
\subsection{Uncertainty-aware Inter-Depth Distillation}
\label{subsec:uncertainty}
As described in~\cite{bts}, the local planar guidance (LPG) layers output LPG depth maps $\{(\mathbf{D}_{\text{lpg}}^{\text{T}})^{i}\}_{i=1}^{3}$ from three scales with the same shape as the input image.
These depth maps provide extra information and result in better performance in the final depth estimation.
In the decoder of the student model, we apply an additional $1\,\times\,1$ convolutional layer to the upsampled feature map of size $\frac{H}{k}\,\times\,\frac{W}{k}\,\times\,C_{k}$, followed by nearest-neighbor interpolation, to produce an intermediate depth map of size $H\,\times\,W\,\times\,1$. We choose $k = 8, 4, 2$ to align with the teacher model, denoting these inter-depth maps as $\{(\mathbf{D}_{\text{inter}}^{\text{S}})^{i}\}_{i=1}^{3}$.



Inspired by the probability density function of the Laplace distribution, we define the uncertainty 
\vspace{-1.5mm}
\begin{equation}
\vspace{-1.5mm}
\label{eq:u}
    \mathbf{U} =  \mathbf{J}_{H,W,1} - \exp(-\frac{|\mathbf{D}_{\text{pred}}-\mathbf{D}_{\text{gt}}|}{\beta |\mathbf{D}_{\text{pred}}+\mathbf{D}_{\text{gt}}|}),
\end{equation}
between the prediction $\mathbf{D}_{\text{pred}}$ and the ground truth $\mathbf{D}_{\text{gt}}$. Here, $\mathbf{J}_{H,W,1}$ denotes a tensor of ones with size $H\,\times\,W\,\times\,1$.
According to Eq. \ref{eq:u}, we consider $\{\mathbf{D}_{\text{inter}}^{\text{S}}\}^{i}$ as prediction and $\{\mathbf{D}_{\text{lpg}}^{\text{T}}\}^{i}$ as the ground truth, yielding uncertainty map $\mathbf{U}^{i}$ for the $i^{\text{th}}$ intermediate depth map.
This uncertainty map indicates how far the student's prediction deviates from the ground truth, serving as a weighting factor to emphasize areas with larger prediction errors. The uncertainty-aware inter-depth distillation loss is then calculated as:
\vspace{-1.5mm}
\begin{equation}
\vspace{-1.5mm}
    L_{\text{KD}}^{\text{D}} = \sum_{i=1}^{3} \frac{1}{2^{i}} || \mathbf{U}^{i} \odot (|(\mathbf{D}_{\text{inter}}^{\text{S}})^{i} - (\mathbf{D}_{\text{lpg}}^{\text{T}})^{i}| ) ||_{1},
\end{equation}
where $\odot$ denotes element-wise multiplication.
\vspace{-0.5mm}

\subsection{Uncertainty-Rectified Depth Loss}
To further guide the depth estimation task, we first accumulate LiDAR point clouds from neighboring frames to generate a dense depth map $\mathbf{D}_{\text{d}}$ as ground truth for training.
However, such aggregation utilizes ego-motion to compensate other frames to the current one, introducing additional errors to densify the point cloud. Authors in~\cite{li2023sparse} also prove that supervising the model with single scan depth, $\mathbf{D}_{\text{s}}$, can enhance depth prediction accuracy. Therefore, we use both $\mathbf{D}_{\text{s}}$ and $\mathbf{D}_{\text{d}}$ to supervise our depth estimation task and focus more on non-zero pixels within the single scan map $\mathbf{D}_{s}$. 
Given the predicted depth $\mathbf{\hat{D}}$, we first calculate the uncertainty maps $\mathbf{U}_{\text{d}}$ and $\mathbf{U}_{\text{s}}$ using Eq. \ref{eq:u}, treating $\mathbf{D}_{\text{d}}$ and $\mathbf{D}_{\text{s}}$ as the ground truth, respectively. These two uncertainty maps are then concatenated and passed through a softmax function along the channel dimension. After this rectification, for each pixel location $p = (x, y)$, we ensure that $\mathbf{U}_{\text{s}}(p) + \mathbf{U}_{\text{d}}(p) = 1$, and additionally decrease the loss weight for the compensated pixel from other frames.

The final uncertainty-rectified depth loss
\vspace{-2.5mm}
\begin{equation}
\vspace{-2.5mm}
\begin{split}
    L_{\text{Depth}} = \frac{1}{|\Omega_{\text{d}}|}\sum_{p\in \Omega_{\text{d}}}||\mathbf{U}_{\text{d}}(p) \odot (|\mathbf{D}_{\text{d}}(p) - \hat{\mathbf{D}}(p)| )||_{1} \\
    + \frac{1}{|\Omega_{\text{s}}|}\sum_{p\in \Omega_{\text{s}}}||\mathbf{U}_{\text{s}}(p) \odot(|\mathbf{D}_{\text{s}}(p) - \hat{\mathbf{D}}(p)|)||_{1},
\end{split}
\vspace{-1.5mm}
\end{equation}
is defined by averaging the weighted differences between the predicted and estimated depth values over valid pixel sets. 
Here, $\Omega_{(\cdot)}$ represent the sets of pixels where $\mathbf{D}_{(\cdot)}$ are valid.

Therefore, the total loss is a combination of multiple loss functions, each with its own weight factor, $\gamma$.
\begin{equation}
    L_{\text{total}} = L_{\text{Depth}} + \gamma_{1}L_{\text{KD}}^{\text{I}}+ \gamma_{2}L_{\text{KD}}^{\text{R}}+ \gamma_{3}L_{\text{KD}}^{\text{Dec}} + \gamma_{4}L_{\text{KD}}^{\text{D}}.
\end{equation}

\label{sec:approach}

\vspace{-3mm}
\section{Experiments}
This section begins by introducing the dataset and implementation details. We then demonstrate the effectiveness of our proposed methods. Finally, we present ablation studies to further highlight the efficiency of our approaches.
\vspace{-0.6mm}
\subsection{Dataset and Implementation Details}
This work employs the nuScenes dataset \cite{nuscenes}, a comprehensive multi-sensor autonomous driving dataset, to train and evaluate the efficiency of our models. To generate the dense training ground truth depth map $\mathbf{D}_{\text{d}}$ from LiDAR, we adopt the accumulation strategy described in \cite{cafnet}. Furthermore, we use the single-frame LiDAR depth map $\mathbf{D}_{\text{s}}$ as an additional supervision to mitigate errors arising from the accumulation process. We retrained the CaFNet \cite{cafnet} using our proposed URDL, resulting in improved performance compared to the original result. Our model is developed using PyTorch \cite{Pytorch} and trained on an Nvidia\textsuperscript{\textregistered} Tesla A30 GPU with a batch size of 6.
The data augmentation techniques, learning rate decay strategies, and evaluation metrics are consistent with \cite{cafnet}.
\vspace{-0.9mm}
\subsection{Quantitative Results}
\begin{table*}[ht]
\vspace{-1mm}
\centering
\caption{Performance Comparison on nuScenes Official Test Set. }
\vspace{-2mm}
\label{tab:exp}
\resizebox{17.5cm}{!} 
{
\centering

\begin{tabular}{c||c|c|c|c|cccccccc}
\hline
\multirow{2}{*}{Eval Distance} & \multirow{2}{*}{Method} & \multicolumn{3}{c|}{Model} & \multicolumn{8}{c}{Metrics} \\ \cline{3-13} 
                                    &                         & Params$\downarrow$      &  FLOPs$\downarrow$ & Runtime$\downarrow$   & MAE $\downarrow$ & RMSE $\downarrow$ & AbsRel $\downarrow$  &log10 $\downarrow$ & RMSElog $\downarrow$ & $\delta_{1}$ $\uparrow$& $\delta_{2}$ $\uparrow$ & $\delta_{3}$ $\uparrow$  \\ \hline
\multirow{4}{*}{50m}                & RadarNet   \cite{radarnet}  & 8.39M+14.41M  & - & 0.336s+0.042s & 1.706 & 3.742 & 0.103 & 0.041 & 0.170 & 0.903 & 0.965 &0.983 \\ 
                                    & CaFNet (T)\textsuperscript{\dag}   & 62.25M &  685G  & 0.132s & 1.359 & 3.131 & 0.079 & 0.031 & 0.145 & 0.931 & 0.974 & 0.987 \\
                                    & LiRCDepth (w/o KD) & 12.65M &  121G & 0.069s & 1.638 & 3.499 & 0.101 & 0.039 & 0.163 & 0.905 & 0.965 & 0.985 \\
                                    & LiRCDepth (KD)      & 12.65M &  121G & 0.069s  & \textbf{1.514} & \textbf{3.330} & \textbf{0.092} & \textbf{0.036} & \textbf{0.155} & \textbf{0.916}  & \textbf{0.969} & \textbf{0.986}\\
                                    \hline
\multirow{4}{*}{70m}                & RadarNet    \cite{radarnet}  & 8.39M+14.41M  & -   & 0.336s+0.042s & 2.073 & 4.591 & 0.105  & 0.043 & 0.181 & 0.896 & 0.962 & 0.981      \\ 
                                    & CaFNet (T)\textsuperscript{\dag} & 62.25M &  685G  & 0.132s &  1.673 & 3.928 & 0.082 & 0.033 & 0.154 & 0.923 & 0.971  & 0.986 \\
                                    & LiRCDepth (w/o KD)  & 12.65M &  121G & 0.069s & 2.037 & 4.479 & 0.104 & 0.042 & 0.174 & 0.894 & 0.960 & 0.982 \\
                                    & LiRCDepth (KD)      & 12.65M &  121G & 0.069s & \textbf{1.898} & \textbf{4.300} & \textbf{0.095} & \textbf{0.036} & \textbf{0.155} & \textbf{0.916}  & \textbf{0.969} & \textbf{0.986}\\
                                    \hline
\multirow{4}{*}{80m}                & RadarNet  \cite{radarnet}  &8.39M+14.41M  & - & 0.336s+0.042s & 2.179 & 4.899 & 0.106 & 0.044 & 0.184 & 0.894 & 0.959 & 0.980 \\ 
                                    & CaFNet (T)\textsuperscript{\dag}  &62.25M &  685G  & 0.132s & 1.763 & 4.184 & 0.083 & 0.034 & 0.156 & 0.921 & 0.970 & 0.985\\
                                    & LiRCDepth (w/o KD) & 12.65M &  121G & 0.069s  & 2.152 & 4.801 & 0.105 & 0.043 & 0.177 & 0.892 & 0.959 & 0.981 \\
                                    & LiRCDepth (KD)      & 12.65M &  121G & 0.069s  & \textbf{2.009} & \textbf{4.617} & \textbf{0.096} & \textbf{0.039} & \textbf{0.170} & \textbf{0.903}  & \textbf{0.963} & \textbf{0.983}\\
                                    \hline
\end{tabular}
}
\\[1pt]
\scriptsize{T: Teacher model, w/o KD: Student model trained without distillation, KD: Student model trained with distillation \textsuperscript{\dag} indicates our reproduced results.}
\vspace{-5mm}
\end{table*}

We evaluate the performance of LiRCDepth on the official nuScenes test set, comparing it against CaFNet \cite{cafnet} and RadarNet \cite{radarnet}. The comparison includes the performance of depth estimation, the number of model parameters, run time, and FLOPs. As shown in Table \ref{tab:exp}, our proposed LiRCDepth model contains 12.65 million parameters, representing approximately an 80\% reduction compared to CaFNet. FLOPs are computed using images with a resolution of $894\,\times\,1600 \times\,3$. Despite the large image size, LiRCDepth achieves a FLOP count of 121G, which is only 20\% of that required by the teacher. Compared to RadarNet, which employs a two-stage training process, LiRCDepth has 10 million fewer parameters and significantly reduced run time while achieving a 7.8\% improvement in Mean Absolute Error (MAE). Furthermore, by applying our proposed distillation strategies, LiRCDepth improves performance by 6.6\% in MAE and 3.8\% in Root Mean Square Error (RMSE) compared to direct training, underscoring the effectiveness of the proposed distillation framework.
\vspace{-0.9mm}
\subsection{Qualitative Results}
\begin{figure}[t]
    \centering
    \includegraphics[width=0.99\linewidth]{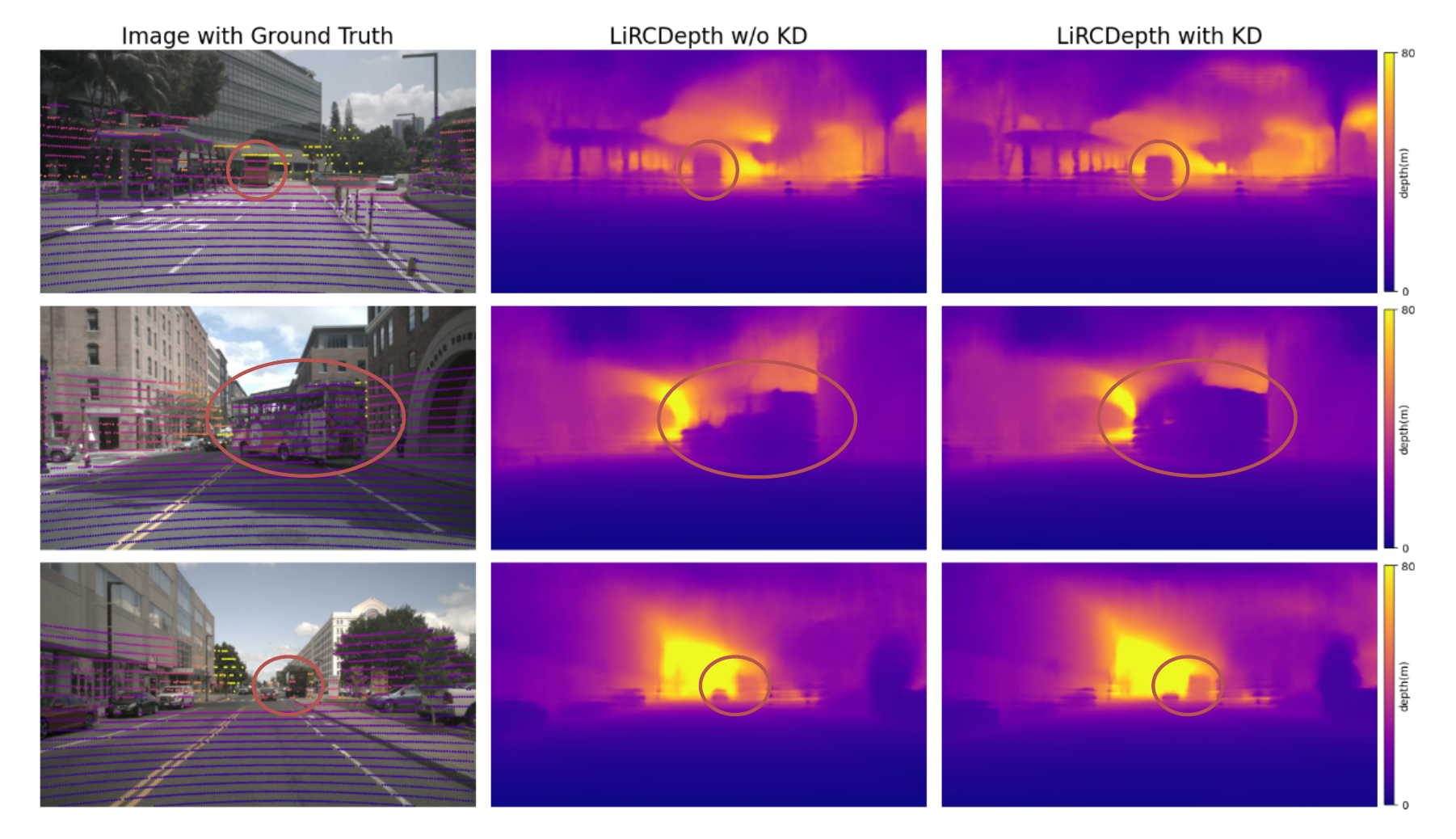}
    
       \caption{Qualitative comparison at 80 meters depth range. Colume 2: LiRCDepth(w/o KD). Colume 3: LiRCDepth(KD).}
    \label{fig:qualitative}
    \vspace{-5mm}
    \end{figure}

As illustrated in Fig. \ref{fig:qualitative}, the model trained with distillation produces depth maps with clearer boundaries. In the first row, LiRCDepth(KD) provides a more defined contour of the bus and the bus station, while in the second row, LiRCDepth(w/o KD) struggles to accurately predict the shape of the bus. The last row demonstrates that LiRCDepth(KD) offers superior predictions in long-range scenarios.
\vspace{-0.5mm}
\subsection{Ablation Studies}
In this section, we present experiments to demonstrate the efficacy of the proposed uncertainty-guided loss. We begin by comparing the performance of the proposed URDL with the conventional L1 loss. Additionally, we conduct further experiments on the depth distillation component by removing the uncertainty $\mathbf{U}$ denoted in Eq. \ref{eq:u} from the distillation loss. The results of these experiments are detailed in Table \ref{table:ablation_depth}.
\setlength\tabcolsep{10pt}
\begin{table}[ht]
\caption{Ablation study on the uncertainty.}
\vspace{-2mm}
\centering
\begin{tabular}{c||cccc}
\hline
Loss & MAE $\downarrow$ & RMSE $\downarrow$ & AbsREL $\downarrow$ & $\delta_{1}$ $\uparrow$\\ \hline
L1 &2.174 & 4.892 & \textbf{0.105}& 0.891 \\
URDL &\textbf{2.152} & \textbf{4.801}& \textbf{0.105}& \textbf{0.892} \\
\hline
$L_{\text{KD}}^{\text{D}}$ w/o $\mathbf{U}$ & 2.072 & 4.703& 0.098& 0.904 \\
$L_{\text{KD}}^{\text{D}}$ & \textbf{2.009}& \textbf{4.617} & \textbf{0.096}& \textbf{0.903}  \\
 \hline
\end{tabular}
\label{table:ablation_depth}
\vspace{-3mm}
\end{table}

Then, we conduct a series of experiments to demonstrate the efficiency of each proposed distillation loss, summarized in Table. \ref{table:ablation_kd}.

\setlength\tabcolsep{5pt}
\begin{table}[ht]
\centering
\vspace{-1mm}
\caption{Ablation study on the distillation losses.}
\vspace{-2mm}
\centering
\begin{tabular}{cccc||cccc}
\hline
$L_{\text{KD}}^{\text{Dec}}$ & $L_{\text{KD}}^{\text{D}}$  & $L_{\text{KD}}^{\text{I}}$ & $ L_{\text{KD}}^{\text{R}}$ & MAE $\downarrow$ & RMSE $\downarrow$ & AbsREL $\downarrow$ & $\delta_{1}$ $\uparrow$\\ \hline
- & - & - & - & 2.152 & 4.801& 0.105& 0.892 \\
\hline
\checkmark & &  &  & 2.119  & 4.752  & 0.103  & 0.894  \\
 &  \checkmark &    &  &  2.145 & 4.781  & 0.105  & 0.894\\
 & & \checkmark &  & 2.128  & 4.753  & 0.103  & 0.893  \\
 & &  &  \checkmark& 2.131  & 4.763  &  0.104 &  0.896 \\
\hline
\checkmark & \checkmark &  &  & 2.093  &  4.701 & 0.100  & 0.901  \\
\checkmark & &\checkmark  &  &  2.087 & 4.692  & 0.099  &  0.901 \\
\checkmark & &  &\checkmark  &  2.091 & 4.702  & 0.099  &  0.899 \\
 & \checkmark& \checkmark &  &  2.093 & 4.710  & 0.100  & 0.900  \\
 & \checkmark&  &  \checkmark& 2.100  & 4.723  & 0.102  &  0.898 \\
 & & \checkmark & \checkmark & 2.108  & 4.731  & 0.101  & 0.899  \\
\hline
\checkmark &\checkmark & \checkmark & & 2.071 & 4.701  &    0.098  & 0.900  \\
\checkmark & & \checkmark & \checkmark &  2.057 & 4.687  & 0.097  & 0.901  \\
 & \checkmark& \checkmark & \checkmark & 2.066  &4.693   & 0.097  &  0.901 \\
\hline
\checkmark &\checkmark & \checkmark & \checkmark &  \textbf{2.009}& \textbf{4.617} & \textbf{0.096}& \textbf{0.903}   \\

 \hline
\end{tabular}
\label{table:ablation_kd}
\vspace{-5mm}
\end{table}
\label{sec:exp}

\section{conclusion}
In this paper, we introduce LiRCDepth, a lightweight radar-camera depth estimation framework that achieves results comparable to other heavy-weight models while utilizing significantly fewer parameters. To train this model effectively, we propose a knowledge distillation framework that focuses on three key aspects, guiding the student model to replicate the performance of the teacher. Additionally, we introduce an uncertainty-rectified depth loss to more accurately guide depth prediction toward the ground truth. Evaluation on the nuScenes dataset demonstrates the effectiveness of our proposed methods. In the future, we plan to extend the work by distilling knowledge from multiple teacher models. 

\label{sec:conclusion}

\section{Acknowledgement}
Research leading to these results 
has received funding from the EU ECSEL Joint Undertaking under grant agreement n° 101007326 (project AI4CSM) and from the partner national funding authorities the German Ministry of Education and Research (BMBF).

\bibliography{mybib} 
\bibliographystyle{IEEEtran} 


\end{document}